\documentclass[default,iicol]{sn-jnl}

\usepackage{graphicx}%
\usepackage{multirow}%
\usepackage{amsmath,amssymb,amsfonts}%
\usepackage{amsthm}%
\usepackage{mathrsfs}%
\usepackage[title]{appendix}%
\usepackage{xcolor}%
\usepackage{textcomp}%
\usepackage{manyfoot}%
\usepackage{booktabs}%
\usepackage{algorithm}%
\usepackage{algorithmicx}%
\usepackage{algpseudocode}%
\usepackage{listings}%
\usepackage{amssymb}
\usepackage{pifont}
\usepackage{bbding}
\usepackage{hyperref}
\usepackage{balance}

\theoremstyle{thmstyleone}%
\theoremstyle{thmstyletwo}%

\theoremstyle{thmstylethree}%

\begin{document}

\title[Article Title]{Pointmap Association and Piecewise-Plane Constraint for Consistent and Compact 3D Gaussian Segmentation Field}


\author[1,2]{\fnm{Wenhao} \sur{Hu}}\email{whu@zju.edu.cn}
\author[3]{\fnm{Wenhao} \sur{Chai}}\email{wchai@uw.edu}
\author[1,2]{\fnm{Shengyu} \sur{Hao}}\email{shengyuhao@zju.edu.cn}
\author[1]{\fnm{Xiaotong} \sur{Cui}}\email{xiaotong.21@intl.zju.edu.cn}
\author[1]{\fnm{Xuexiang} \sur{Wen}}\email{xuexiangwen@zju.edu.cn}
\author[3]{\fnm{Jenq-Neng} \sur{Hwang}}\email{hwang@uw.edu}
\author*[1,2]{\fnm{Gaoang} \sur{Wang}}\email{gaoangwang@intl.zju.edu.cn}

\affil[1]{\orgdiv{ZJU-UIUC Institute}, \orgname{Zhejiang University}, \orgaddress{ \street{718 East Haizhou Rd.}, \city{Haining}, \postcode{314400}, \state{Zhejiang}, \country{China}}}
\affil[2]{\orgdiv{College of Computer Science and Technology}, \orgname{Zhejiang University}, \orgaddress{ \street{38 Zheda Rd.}, \city{Hangzhou}, \postcode{310027}, \state{Zhejiang}, \country{China}}}
\affil[3]{\orgdiv{Department of Electrical \& Computer Engineering}, \orgname{University of Washington}, \orgaddress{\street{185 Stevens Way}, \city{Seattle}, \postcode{98195}, \state{Washington}, \country{USA}}}


 \abstract{
Achieving a consistent and compact 3D segmentation field is crucial for maintaining semantic coherence across views and accurately representing scene structures. 
Previous 3D scene segmentation methods rely on video segmentation models to address inconsistencies across views, but the absence of spatial information often leads to object misassociation when object temporarily disappear and reappear. 
Furthermore, in the process of 3D scene reconstruction, segmentation and optimization are often treated as separate tasks. As a result,  optimization typically lacks awareness of semantic category information, which can result in floaters with ambiguous segmentation.
To address these challenges, we introduce CCGS, a method designed to achieve both view \underline{C}onsistent 2D segmentation and a \underline{C}ompact 3D \underline{G}aussian \underline{S}egmentation field. CCGS incorporates pointmap association and a piecewise-plane constraint.
First, we establish pixel correspondence between adjacent images by minimizing the Euclidean distance between their pointmaps. We then redefine object mask overlap accordingly. The Hungarian algorithm is employed to optimize mask association by minimizing the total matching cost, while allowing for partial matches.
To further enhance compactness, the piecewise-plane constraint restricts point displacement within local planes during optimization, thereby preserving structural integrity.
Experimental results on ScanNet and Replica datasets demonstrate that CCGS outperforms existing methods in both 2D panoptic segmentation and 3D Gaussian segmentation.  \href{https://whuechoscript.github.io/CCGS}{Project Page}
}




\maketitle
\section{Introduction}\label{sec1}
3D scene segmentation plays a crucial role in enhancing scene perception, understanding, and interaction~\cite{rozenberszki2024unscene3d,huang2024segment3d,liu2024active,castillo2024contrastive,xu2024embodiedsam}. It facilitates diverse and advanced applications such as autonomous driving, augmented reality, and robotics by enabling efficient navigation, precise object recognition, and intelligent interaction within complex environments~\cite{kamran2024applications,kong2024multi,kong2023robo3d,xia2023scpnet,bian2024dynamiccity}. Recent advancements have extensively explored NeRF-based segmentation methods~\cite{ bhalgat2023contrastive, siddiqui2023panoptic,zhi2021place,wang2022dm}, which allow for segmenting and rendering of images from novel viewpoints. However, these methods often suffer from slow training and inference times, limiting their practicality in real-world scenarios. The emergence of 
Gaussian splatting~\cite{kerbl20233d} introduces a innovative approach that enables real-time 2D rendering while maintaining distinct spatial meaning for each Gaussian point. This advancement offers a new perspective on constructing 3D segmentation fields, addressing limitations of earlier methods.

\begin{figure}[t]
    \centering
    \includegraphics[width=1\linewidth]{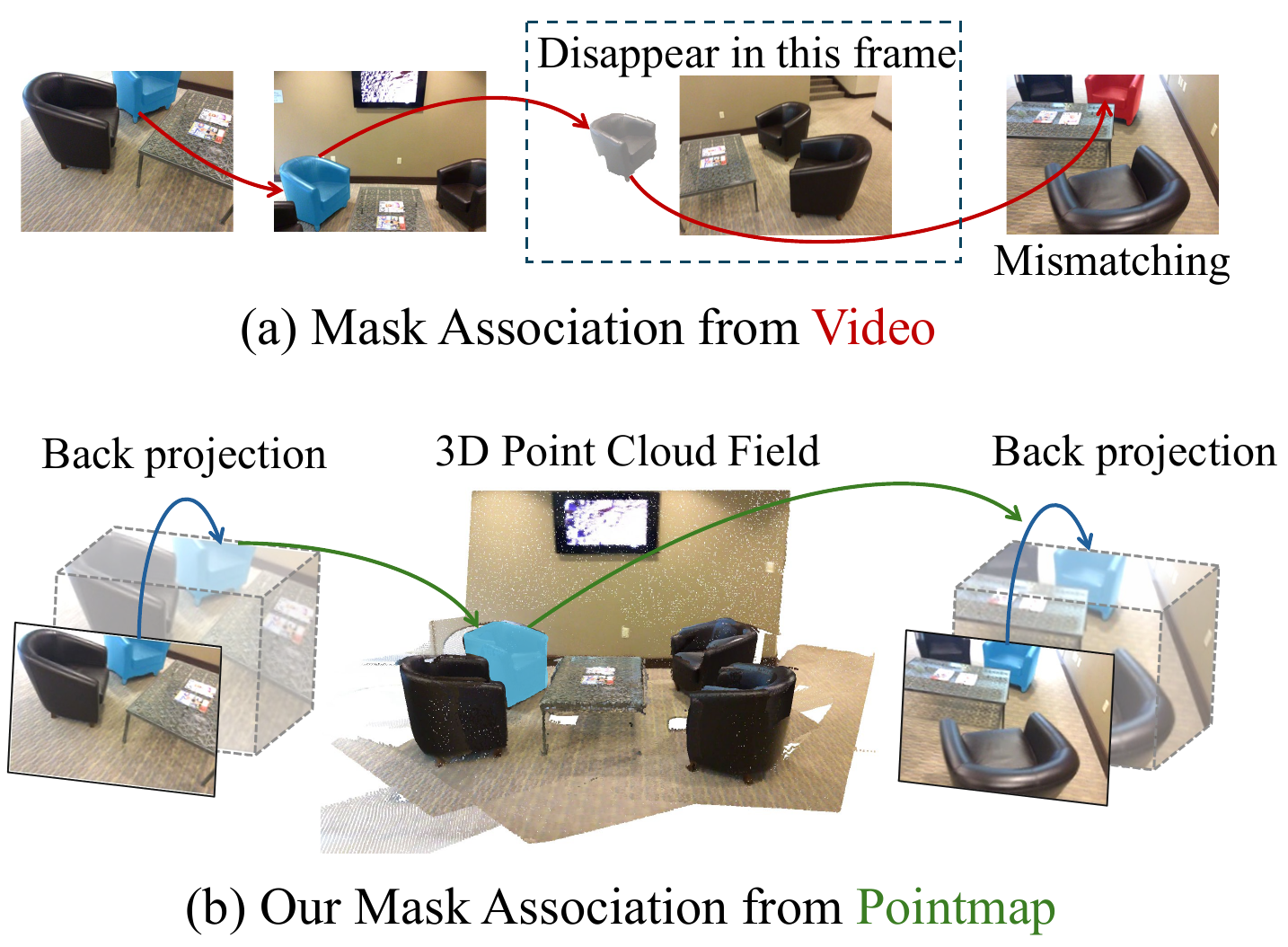}
    \caption{Differences in mask association: Video vs. Pointmap. Video segmentation often struggle to maintain consistency during significant changes in camera views. In contrast, constructing a unified 3D point cloud field can ensure segmentation accuracy by leveraging spatial information.}
    \label{fig1}
\end{figure}

Previous research~\cite{dou2024cosseggaussians,hu2024semantic,wu2024opengaussian, ye2023gaussian,zhou2024feature,zanjani2024planar} highlights two main challenges in establishing a 3D Gaussian segmentation field. First, some existing methods~\cite{dou2024cosseggaussians,ye2023gaussian} rely on video segmentation models~\cite{cheng2023tracking} for mask association, which lack spatial consistency. This limitation becomes evident in complex scenes where objects intermittently disappear and reappear in the image sequence due to occlusion or moving out of view. In such cases, the same object may be assigned different IDs, reducing the effectiveness of these methods. For instance, as shown in Figure~\ref{fig1}, when an object like a chair exits the frame and reappears several frames later, the video segmentation model may mistakenly assign it a new ID, treating it as a different object. 
Second, segmentation and reconstruction are typically handled as separate tasks, leading to an optimization process that lacks semantic awareness and fails to incorporate meaningful structural information. 
The absence of effective semantic constraints during the optimization process often leads to the generation of meaningless floaters in empty space, which do not contribute to the segmentation field. Furthermore, the lack of constraints during the densification process allows replicated points to spread without restriction, resulting in ambiguous class boundaries. These limitations compromise both the accuracy and compactness of the 3D Gaussian segmentation field, making it less effective in representing complex scenes.

To address the aforementioned issues, we propose CCGS, a method combining pointmap-based association and a piecewise-plane constraint to construct a consistent and compact 3D Gaussian segmentation field. This approach ensures the consistency of 2D segmentation while maintaining the compactness of the 3D segmentation. 
Specifically, we first determine the pixel correspondence between adjacent images by minimizing the Euclidean distance between their pointmaps. The overlap between object masks is redefined based on this correspondence, and a matching cost is computed to associate masks across frames. The mask association problem is solved using the Hungarian algorithm, optimizing the total matching cost while allowing partial matches.
To achieve a compact 3D segmentation field, we introduce plane constrained Gaussian Splatting, where each point is restricted to a piecewise-plane defined by its nearest neighbors of the same class. This constraint enhances the compactness of the segmentation field during optimization. Additionally, in the densification process, we propose a split projection method that ensures replicated points remain within the neighborhood plane of similar points, preventing boundary blending between different classes and preserving segmentation accuracy.

In summary, our work makes the following contributions: 
\begin{itemize}
    \item We propose CCGS, a Gaussian segmentation field that achieves both view-consistent 2D segmentation and compact 3D segmentation.
    \item We propose pointmap association to generate unified 3D field that facilitates  consistent 2D segmentation.
    \item We define piecewise-plane constrained Gaussian splatting, which restricts point displacement during optimization and densification to achieve a compact 3D segmentation.  
    \item We conducted extensive experiments on multiple datasets, demonstrating that our approach achieves state-of-the-art (SOTA) performance in both 2D and 3D segmentation tasks.
\end{itemize}

\section{Related Works}
\subsection{Point Cloud Segmentation}
3D point cloud segmentation classifies the point cloud into meaningful regions or segments that belong to the same class~\cite{tang2022contrastive,park2022fast,peng2023openscene}. Some 3D point cloud segmentation methods primarily rely on training closed-set models. PointNet~\cite{qi2017pointnet} directly learns a spatial encoding of each point, and PointNet++~\cite{qi2017pointnet++} extends it with a local feature extractor based on Farthest Point Sampling (FPS) and is trained with hierarchical feature learning architecture. There are other methods initially processing 2D images and then mapping the segmented 2D results onto the corresponding 3D coordinates of the point cloud. MVPNet~\cite{jaritz2019multi} aggregates 2D multi-view image features into 3D point clouds, and then uses point-based networks to fuse features in 3D canonical space to predict 3D semantic labels. VMVF~\cite{kundu2020virtual} selects various virtual views to render multiple 2D channels for training an effective 2D semantic segmentation model and then fuses features from these predictions onto the 3D mesh vertices to determine semantic labels. However, these methods depend on existing point cloud data as input, which limits their versatility for downstream tasks.
\subsection{Nerf Segmentation}
Semantic-NeRF~\cite{zhi2021place} was the first to integrate semantics into NeRF by fusing noisy 2D segmentations into a consistent 3D model, improving segmentation accuracy and enabling novel view synthesis. Panoptic NeRF~\cite{fu2022panoptic} and DM-NeRF~\cite{wang2022dm} explore panoptic radiance fields for label transfer and scene editing, but both rely on manual ground truth annotations. Panoptic Neural Fields~\cite{kundu2022panoptic} decomposes scenes into objects and backgrounds using instance-specific MLPs for objects and a shared MLP for the background, optimized jointly from color images and predicted segmentations.
Several studies~\cite{kerr2023lerf,liu2023weakly} have explored the approach of lifting latent features from 2D foundation models~\cite{radford2021learning} into 3D space to enable open-vocabulary text queries. Other approaches~\cite{shen2023distilled,goel2023interactive,wei2024nto3d,kim2024garfield} have demonstrated promising results in object-level segmentation tasks. Panoptic Lifting~\cite{siddiqui2023panoptic} and Contrastive Lift~\cite{bhalgat2023contrastive} generate 3D panoptic representations by lifting 2D machine-generated segmentation masks to 3D for multi-view consistency. While Panoptic Lifting~\cite{siddiqui2023panoptic} addresses inconsistencies in 2D instance identifiers through linear assignment, Contrastive Lift~\cite{bhalgat2023contrastive} uses contrastive clustering. Despite their success in multi-view consistent segmentation, NeRF-based methods are limited by slow rendering speeds and high memory usage during training due to their implicit nature.

\subsection{Gaussian Segmentation}
Segmenting the Gaussian field involves dividing it into distinct regions based on their properties, which is crucial for scene reconstruction and understanding. LangSplat~\cite{qin2024langsplat}, LEGaussians~\cite{shi2024language} and several works~\cite{chen2024ovgaussian,peng2024gags,liang2024supergseg,hu2024sparselgs,cheng2024occam,peng20243d,qiu2024gls} incorporate language features from CLIP for open-world scene representation. SADG~\cite{li2024sadg} specifically targets segmentation in dynamic scenes. For single-object segmentation, GaussianCut~\cite{jain2024gaussiancut} proposes a Gaussian distribution-based optimization framework, while GradiSeg~\cite{li2024gradiseg} develops a novel gradient-driven segmentation approach. PLGS~\cite{wang2024plgs} adopts a methodology similar to Panoptic Lifting~\cite{siddiqui2023panoptic}. InstanceGaussian~\cite{li2024instancegaussian} and BCG~\cite{zhang2024bootstraping} propose clustering-based methods for segmenting 3D Gaussian representations.
SAGA~\cite{cen2023segment} efficiently embeds 2D segmentation features into 3D Gaussian point features using contrastive learning. Feature 3DGS~\cite{zhou2024feature} enables 3D Gaussian splatting on semantic features via 2D foundation model distillation to extract arbitrary-dimension semantic features. Gaussian Grouping~\cite{ye2023gaussian} and CoSSegGaussians~\cite{dou2024cosseggaussians} apply video segmentation methods to unify segmentation IDs from multiple views, However, video segmentation methods often fail when there are significant changes in viewing angles. OpenGaussian~\cite{wu2024opengaussian} achieves consistent 3D segmentation through codebook discretization but cannot render precise 2D segmentations. Gaga~\cite{lyu2024gaga} shares a similar motivation with our work. However, it does  not explicitly consider the role of segmentation in Gaussian optimization, and the lack of segmentation-constrained optimization can result in meaningless floaters in the 3D segmentation field. Our approach overcomes these challenges by using point map fusion and plane-constrained Gaussian splatting to create a compact and consistent 3D Gaussian segmentation field.

\section{Method}
\begin{figure*}[t!]
    \centering
    \includegraphics[width=1\linewidth]{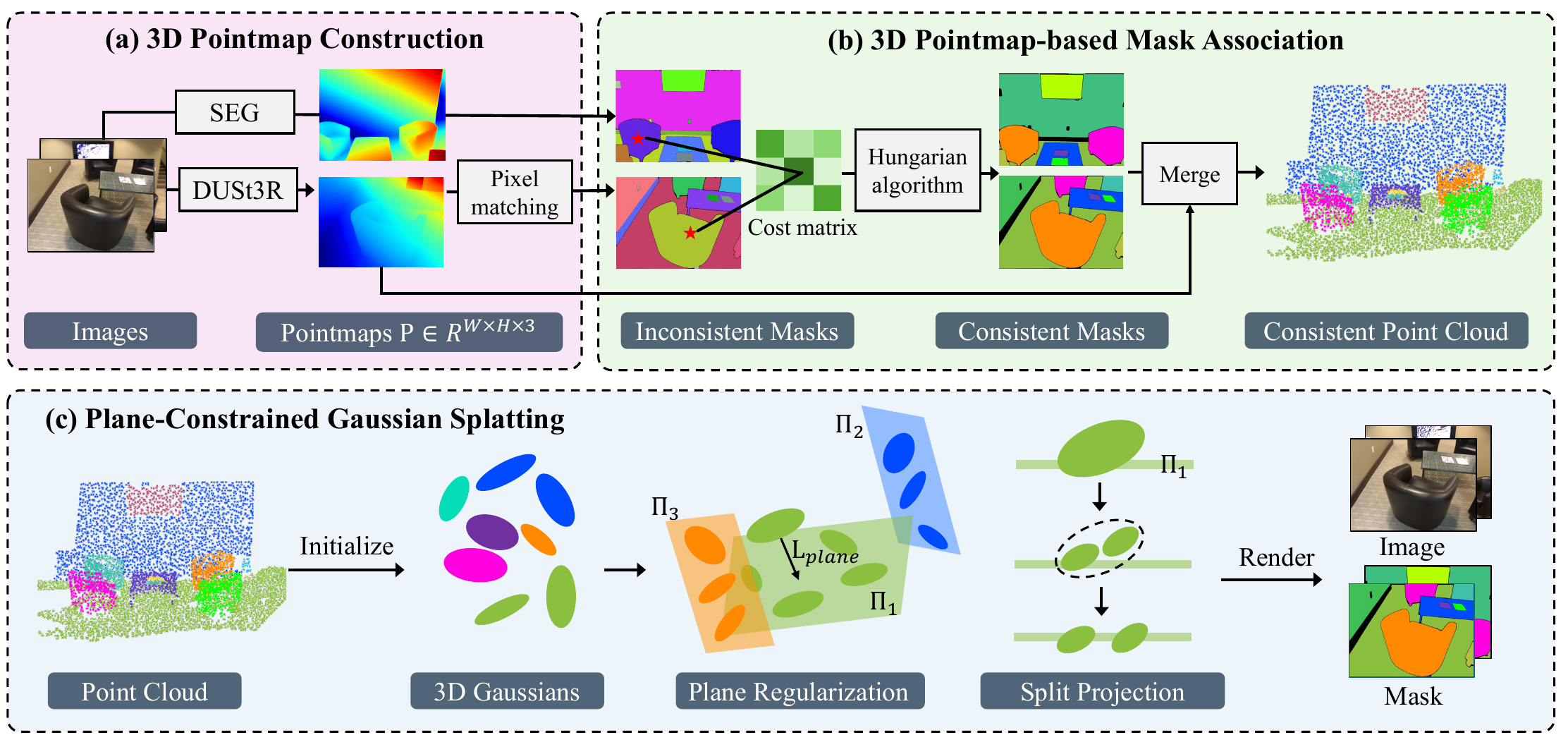}
    \caption{
    The pipeline of our method. (a) We first construct a unified point cloud field and establish correspondences between pixels using the pointmaps. (b) Leveraging these relationships, we construct a cost matrix for instance masks across two frames. The Hungarian algorithm is then applied to optimize the cost matrix, ensuring consistent mask association. By merging all frames, we obtain a point cloud enriched with consistent segmentation information. (c) This point cloud serves as the initialization for 3D Gaussians. To achieve compact 3D segmentation, we employ a piecewise-plane constraint, restricting point displacement within local planes through plane regularization and split projection.}
    \label{framework}
\end{figure*}

\subsection{Preliminaries}\label{3DGS}
3D Gaussian splatting~\cite{kerbl20233d} presents a powerful approach to scene representation, offering impressive reconstruction quality and faster rendering speeds compared to methods like Neural Radiance Field~\cite{mildenhall2021nerf}. It achieves this by utilizing 3D Gaussians to explicitly represent scene geometry and appearance, resembling a point cloud. Each Gaussian is defined by its centroid $\boldsymbol{x}$, 3D covariance $\boldsymbol{\Sigma}$, opacity $\boldsymbol{\alpha}$, and color $\boldsymbol{c}$, represented as spherical harmonics (SH) coefficients of three degrees.

To effectively supervise these learnable attributes, Gaussian splatting projects them onto the 2D imaging plane for rendering RGB images from a given viewpoint. This projection is performed via $\alpha$-blending, a differentiable rasterization method optimized for GPU implementation. For each pixel, the color $\textit{C}$ 
 is computed as follows:
\begin{equation}
    \textit{C} = \sum_{i \in \mathcal{N}} \boldsymbol{c}_i \boldsymbol{\alpha}_i' \prod_{j=1}^{i-1}(1 - \boldsymbol{\alpha}_j')
\end{equation}
Here, $\boldsymbol{c}_i$ represents the color of the $i$-th Gaussian, $\boldsymbol{\alpha}_i'$ denotes its influence factor calculated by multiplying the projected 2D covariance with the per-point opacity $\boldsymbol{\alpha}_i$.

To extend the Gaussian Rasterizer from color space $\textit{C}$ to segmentation space $\textit{S}$, Identity Encoding $\boldsymbol{s}_i$~\cite{ye2023gaussian} is introduced. The Identity Encoding is a learnable vector of length 16 that represents the segmentation feature for each Gaussian, enabling precise and differentiable instance segmentation within the Gaussian framework. For each pixel, the segmentation $\textit{S}$ can be expressed as: 
\begin{equation}
\label{eq:S}
\textit{S} = \sum_{i \in \mathcal{N}} \boldsymbol{s}_i \boldsymbol{\alpha}_i '\prod_{j=1}^{i-1} (1 - \boldsymbol{\alpha}_j')
\end{equation}

\subsection{Pointmap-based Association for Consistent Segmentation}
Mask association methods in video segmentation models often struggle with complex scenes involving multiple objects. To address these challenges, we propose a pointmap-based association approach to create a 3D segmentation field that aligns instance IDs across different viewpoints. By incorporating spatial information, our method achieves consistent and reliable multi-view segmentation.

The process begins with the creation of a unified 3D pointmap field. A pointmap \( P \in \mathbb{R}^{W \times H \times 3} \) represents a dense 2D field of 3D points, establishing a precise one-to-one correspondence between the pixels of an RGB image \( I \) with resolution \( W \times H \) and their respective 3D scene points. To achieve this, we employ DUSt3R~\cite{wang2023dust3r} to construct a unified pointmap field \( \mathcal{P} \) that integrates individual pointmaps \( \{P_1, P_2, \ldots, P_n\} \). Subsequently, the images  \( \{I_1, I_2, \ldots, I_n\} \) are processed by a segmentation model~\cite{kirillov2023segment} to generate a set of inconsistent masks \(\{M_1, M_2, \ldots, M_n\}\).

Given two adjacent images, \( I_t \) and \( I_{t+1} \), along with their corresponding pointmaps, \( P_t \) and \( P_{t+1} \), we establish pixel correspondence by measuring the distance between the pointmaps. For each pixel \( (i, j) \) in image \( I_t \), we identify the closest pixel \( (k, l) \) in image \( I_{t+1} \) by minimizing the Euclidean distance between their respective pointmap values:
\begin{equation}
(k, l) = \arg\min_{(k', l')} \| P_t(i, j) - P_{t+1}(k', l') \|_2
\end{equation}
This correspondence is formally represented as a mapping function \( \phi: (i, j) \mapsto (k, l) \), where  
\begin{equation}
\phi(i, j) = (k, l)
\end{equation}

For \( a^{th} \) object mask in \( M_{t} \) and \( b^{th} \) object mask in \( M_{t+1} \),  
we redefine the overlap between two object masks \( M_{t}^a \) and \( M_{t+1}^b \) based on \( \phi \):  

\begin{equation}
\begin{split}
M_{t}^a \cap_\phi M_{t+1}^b &= \left\{ (i, j) \mid (i, j) \in M_{t}^a , \right. \\
&\quad \left. \phi(i, j) \in  M_{t+1}^b  \right\}
\end{split}
\end{equation}
This means that a pixel \( (i, j) \) contributes to the intersection only if it belongs to \( M_{t}^a \) and its corresponding pixel \( (k, l) = \phi(i, j) \) belongs to \( M_{t+1}^b \). Denote \( \lvert \cdot \rvert \) as the number of points in a mask. Using this refined intersection definition, the matching cost between classes \( a \) and \( b \) is given by:  
\begin{equation}
C(a, b) = 1 - \frac{|M_{t}^a \cap_\phi M_{t+1}^b|}{\min(|M_{t}^a|, |M_{t+1}^b|)}
\end{equation}

The mask association problem can be effectively addressed using the Hungarian algorithm, which aims to minimize the total matching cost between objects in \( M_t \) and \( M_{t+1} \). Specifically, let \( \mathbb{I}_{a,b} \in \{0, 1\} \) be a binary variable that indicates whether the object \( M_t^a \) in the current frame is matched with the object \( M_{t+1}^b \) in the next frame. The objective is to determine the optimal segmentation correspondence function \( \psi: \mathcal{A} \mapsto \mathcal{B} \), which assigns object in \( M_t \) to its most appropriate counterpart in \( M_{t+1} \) by minimizing the total association cost. The objective function for this optimization problem is defined as:
\begin{equation}
\psi = \arg\min_{\psi} \sum_{a \in \mathcal{A}} \sum_{b \in \mathcal{B}} C(a, b) \cdot \mathbb{I}_{a,b}
\end{equation}
The constraints are relaxed to allow partial matching: 
\begin{equation}
\begin{cases}
\sum_{b \in \mathcal{B}} \mathbb{I}_{a,b} \leq 1, & \forall a \in \mathcal{A} \\
\sum_{a \in \mathcal{A}} \mathbb{I}_{a,b} \leq 1, & \forall b \in \mathcal{B} \\
\mathbb{I}_{a,b} \in \{0, 1\}, & \forall a \in \mathcal{A}, \forall b \in \mathcal{B}
\end{cases}
\end{equation}

Through this matching process, we obtain a set of consistently labeled 2D masks, \( \boldsymbol{m} \), for all multi-view images, along with a 3D segmented point cloud field, \( \mathcal{P}^s \). These unified 2D masks serve as pseudo-labels for the subsequent training of the Gaussian field, ensuring cross-view consistency and alignment. Meanwhile, the 3D segmented point cloud field \( \mathcal{P}^s \), constructed by progressively integrating associated points, is utilized to initialize the Gaussian splatting process.

\subsection{Piecewise-Plane Constrained Gaussian Splatting}
To ensure compact 3D segmentation during training, we introduce a piecewise-plane constraint. For each point position \(\boldsymbol{x}_i\) in the Gaussian field \(\mathcal{G}\), its nearest neighbors \(\mathcal{N}(\boldsymbol{x}_i)\) of the same class are used to fit a local plane \(\Pi_i\). The constraint is then formulated to minimize the distance between each point \(\boldsymbol{x}_i\) and its corresponding local plane \(\Pi_i\). By enforcing this constraint, points are encouraged to remain close to their class-specific neighborhood planes during optimization and densification, thus maintaining the structural integrity and compactness of the 3D segmentation. This approach provides two key benefits: first, the plane constraint during optimization reduces the occurrence of floating points, ensuring that all points adhere to the object's surface, thereby preserving the compactness of the 3D segmentation field. Second, during densification, replicated points are restricted to the local neighborhood of similar points, minimizing ambiguity and misclassification at object boundaries. To simplify computation, the piecewise-planes are updated every 1000 iterations. 

\paragraph{Plane Regularization}
To implement the piecewise-plane constraint, we minimize the distance of points to their corresponding planes. For a given point position \(\boldsymbol{x}_i\), let \((a_i, b_i, c_i)\) represent the coefficients of the normal vector \(\boldsymbol{n}_i\) of its neighborhood plane. The equation of the neighborhood plane \(\Pi_i\) can be expressed as:

\begin{equation}
\Pi_i: a_i x + b_i y + c_i z + D_i = 0
\end{equation}
An arbitrary point \(\boldsymbol{x}_p\) lying on the plane \(\Pi_i\) is used to compute the perpendicular distance from \(\boldsymbol{x}_i\) to the plane. The distance is calculated as:

\begin{equation}
d_i = \left| \boldsymbol{n}_i^T \cdot (\boldsymbol{x}_i - \boldsymbol{x}_p) \right|
\end{equation}
To enforce this constraint across the entire 3D Gaussian field, we define a plane regularization loss as the average distance of all points to their respective planes. Given \(r\) Gaussian points, the loss is expressed as:

\begin{equation}
\mathcal{L}_\text{plane} = \frac{1}{r} \sum_{i=1}^r d_i.
\end{equation}

As illustrated in Figure~\ref{framework}, incorporating this plane regularization loss ensures that points are primarily adjusted within their respective categories, improving the coherence and precision of the segmentation and reconstruction process.

\paragraph{Split projection}
The adaptive control process identifies points with excessively large gradients in the Gaussian position $\boldsymbol{x}$ for densification. At these points, smaller Gaussians are cloned, while larger Gaussians are split. 
During the cloning or splitting process of Gaussians, we replicate their original class assignments to ensure that points copied from one class are not optimized as belonging to another. This prevents boundary confusion among Gaussians. Ultimately, by following the same-class cloning rule, we obtain \( \mathcal{P}^{s'} \) from the initial segmentation field \( \mathcal{P}^s \).
We denote the cloned Gaussians position as $\boldsymbol{x}^c$ and the split Gaussians position as $\boldsymbol{x}^s$. The cloned Gaussians $\boldsymbol{x}^c$ replicate the original position and shifts in the direction of the positional gradient, with plane optimization constraining its new location.
For split Gaussians \( \boldsymbol{x}^s \), the positions of the newly generated points are determined by sampling from the original 3D Gaussian distribution. The original 3D Gaussian is treated as a probability density function, guiding the placement of these new points. However, this sampling-based initialization may cause the new points to deviate from the intended piecewise-plane structure. Consequently, different classes tend to mix at the boundaries after splitting.
To prevent split Gaussians from being optimized as other categories, we project the split points onto the piecewise-plane. If $\boldsymbol{x}^s_i$ is the point to be projected, the vertical distance from the point to the plane is $d_i$, and the split projection operation, which maps the point \(\boldsymbol{x}^s_i\) onto the plane, can be defined as:
\begin{equation}
 \boldsymbol{x}^{s\prime}_i = \boldsymbol{x}^s_i - d_i \boldsymbol{n}_i
\end{equation}

\definecolor{global}{RGB}{21,96,130}
\definecolor{breakpoint}{RGB}{51,0,111}

\algnewcommand{\Comment}[1]{\textcolor{blue}{\(\triangleright\) #1}}
\begin{algorithm}[h!]
\caption{Pseudocode for CCGS}\label{alg:optimization}
\begin{algorithmic}
\Require Input images $\mathcal{I} =  \{I_1, I_2, \ldots, I_n\}$, Consistent masks $\boldsymbol{m} = \emptyset $, Segmented point cloud $\mathcal{P}^s = \emptyset $
\State \textcolor{global}{\# Get inconsistent segmentation}
\State $\{M_1, M_2, \ldots, M_n\} \gets \text{SEG}(\mathcal{I} )$
\State \textcolor{global}{\# Get Pointmap}
\State $\{P_1, P_2, \ldots, P_n\} \gets \text{DUSt3R}(\mathcal{I} )$
\State $m_1 \gets M_1$
\For{$t \in \{1, 2, \dots, n-1\}$}
    \State \textcolor{global}{\# Establish pixel correspondence}
    \For{$(i, j) \in I_t$}
        \State $\phi(i, j) \gets \arg\min \| P_t(i, j) - P_{t+1}(k', l') \|_2$
    \EndFor
    \State \textcolor{global}{\# Compute mask overlap and cost}
    \State $C(a, b) \gets 1 - \frac{|M_t^a \cap_\phi M_{t+1}^b|}{\min(|M_t^a|, |M_{t+1}^b|)}$
    \State \textcolor{global}{\# Solve mask association}
    \State $\psi \gets \text{Hungarian}(C)$
    \State $m_{t+1} \gets M_{t+1}^{\psi(m_t)} $
     \State \textcolor{global}{\# Get segmented point cloud}
    \State $P^s_{t+1} = \text{cat}(m_{t+1},P_{t+1})$
    \State $\boldsymbol{m} \gets \boldsymbol{m}\cup m_{t+1}$, $\mathcal{P}^s \gets \mathcal{P}^s\cup P^s_{t+1}$
   
\EndFor

\State \textcolor{global}{\# Piecewise-plane constrained gaussian splatting }
\While{not converged}
			\If{IsUpdataPlaneIteration}
                \State $\Pi \gets$ CalculatePiecewise-Plane($\boldsymbol{x}, \boldsymbol{s}$)
                \EndIf	
                
			\State $\hat{I}, \hat{\boldsymbol{m}} \gets$ Rasterize($\boldsymbol{x}$, $\boldsymbol{s}$, $\boldsymbol{\Sigma}$, $\boldsymbol{c}$, $\boldsymbol{\alpha}$)	
                \State $\mathcal{L}_\text{img} \gets \mathcal{L}(I, \hat{I})$
                \State $\mathcal{L}_\text{2d} \gets \mathcal{L}(\boldsymbol{m}, \hat{\boldsymbol{m}})$, $\mathcal{L}_\text{3d} \gets \mathcal{L}(\mathcal{P}^{s}, \boldsymbol{s})$
                \State \textcolor{global}{\# Plane Regularization}
                \State $\mathcal{L}_\text{plane} \gets \mathcal{L}(\Pi, \boldsymbol{x})$
			\State $\mathcal{L}_\text{render}\gets \mathcal{L}_\text{img} + \mathcal{L}_\text{2d} + \mathcal{L}_\text{3d} +\mathcal{L}_\text{plane}$
    
			\State $\boldsymbol{x}, \boldsymbol{s}, \boldsymbol{\Sigma}, \boldsymbol{c}, \boldsymbol{\alpha}$ $\gets$ Adam($\nabla \mathcal{L}_\text{render}$) 
			
			\If{IsRefinementIteration}
                \State \textcolor{global}{\# Densification}
                \State $\boldsymbol{x}^s, \boldsymbol{x}^c \gets$ 
                D($\boldsymbol{x}$)
                \State \textcolor{global}{\# Split Projection}
			\State $\boldsymbol{x}^s \gets$ 
                $\boldsymbol{x}^s - d \boldsymbol{n}$
               
			\EndIf
			\EndWhile
\label{alg}
\end{algorithmic}
\end{algorithm}

\paragraph{Training objective}\label{loss}
Similar to~\cite{ye2023gaussian}, we use a linear layer $f$ followed by a softmax function to map the rendered 2D features $\textit{S}$ in Equation \eqref{eq:S} to a $\mathcal{K}$ classification space. For this 2D classification, we employ a standard cross-entropy loss $\mathcal{L}_{2d}$, which ensures accurate pixel-level mask predictions.
\begin{equation}
    \mathcal{L}_\text{2d} = - \sum_{k \in \mathcal{K}} \boldsymbol{m}[k] \log (softmax(f(\textit{S}))[k])
\end{equation}

To maintain the consistency of 3D segmentation, we also imposed cross-entropy constraints $\mathcal{L}_{3d}$ on the 3D segmentation features $s$ of each point, using the 3D segmentation merged from the pointmap as a pseudo label. 
\begin{equation}
    \mathcal{L}_\text{3d} = - \sum_{k \in \mathcal{K}} \mathcal{P}^{s}[k] \log (softmax(f(\textit{s}))[k])
\end{equation}

Combined with the conventional 3D Gaussian Loss $\mathcal{L}_\text{img}
= (1 - \lambda) \mathcal{L}_1 + \lambda \mathcal{L}_{\text{D-SSIM}}$ on image rendering~\cite{kerbl20233d}, the total loss $\mathcal{L}_\text{render}$ for fully end-to-end training is:
\begin{equation}
    \mathcal{L}_\text{render}  = \mathcal{L}_\text{img}
    + \lambda_\text{plane}\mathcal{L}_\text{plane}
    + \lambda_\text{2d}\mathcal{L}_\text{2d} 
    + \lambda_\text{3d}\mathcal{L}_\text{3d} 
\end{equation}
The final pseudocode is presented in Algorithm~\ref{alg:optimization}.

\section{Experiment}
\subsection{Experimental Setup}

\paragraph{Datasets}
We present experimental results on two datasets: Replica~\cite{straub2019replica} and ScanNet~\cite{dai2017scannet}. The Replica Dataset consists of high-quality reconstructions of various indoor scenes, with each scene containing RGB images paired with corresponding 2D segmentation masks. ScanNet is a large-scale real-world dataset, where each scene comprises images accompanied by annotated 2D segmentation masks. For both datasets, we select 7 scenes for training and evaluation, following a similar approach to~\cite{dou2024cosseggaussians}. Each scene contains approximately 200 training images and 50 testing images, which are uniformly sampled from the dataset. To generate training labels, we utilize 
SAM~\cite{kirillov2023segment}, which produces high-quality object masks.
The annotated segmentation masks provided by the dataset are only available during evaluation as ground-truth labels and are not used during training.

\paragraph{Data availability statement}
The Replica dataset (\url{https://doi.org/10.48550/arXiv.1906.05797}) and can be accessed at \url{https://github.com/facebookresearch/Replica-Dataset}. We use pre-rendered Replica dataset provided by Semantic-NeRF (\url{https://doi.org/10.48550/arXiv.2103.15875}). The ScanNet dataset (\url{https://doi.org/10.48550/arXiv.1702.04405}) and can be accessed at \url{http://www.scan-net.org}.

\paragraph{Evaluation metrics}
For 2D segmentation, we use mean intersection over union (mIoU) to evaluate the quality of predicted masks. In a single-view setting, $mIoU_{s}$ is computed by averaging IoU values across all predicted and ground-truth masks, with optimal assignments determined via linear sum assignment. For multi-view segmentation, we construct a global IoU matrix by aggregating IoU values of masks with the same ID across different viewpoints. The final $mIoU_{m}$ is computed from these aggregated values, providing a unified evaluation across multiple views. Additionally, We calculate PSNR and SSIM to evaluate the quality of rendered images. 
For 3D Gaussian segmentation, we abstract Gaussians into a segmentation field composed of spatial coordinates \( x \) and segmentation features \( s \). To evaluate the segmentation quality, we first align the ground truth point cloud segmentation and reconstructed Gaussian fields to the same scale and orientation. The ground truth labels are then mapped onto the reconstructed field using a nearest neighbor approach. If the nearest distance exceeds a threshold \( \gamma \), the point is assigned as ‘no category’, introducing a penalty for floaters in free space. We then compute the $mIoU_{3D}$ on the reconstructed segmentation field as the primary evaluation metric. Additionally, to further assess the geometric fidelity of the reconstruction, we employ Chamfer Distance to quantify the structural similarity between the ground truth and the reconstructed Gaussian field.

\paragraph{Implementation details}
We implement our method based on Gaussian Grouping~\cite{ye2023gaussian}. For a comprehensive and fair comparison, our method, along with Gaussian Grouping and OpenGaussian, operates on the same initial point cloud for 3D Gaussian segmentation and various downstream tasks. The threshold parameters, $\gamma$ is set to 0.5. During training, we set $\lambda_\text{plane} = 10$, $\lambda_\text{2d} = 1$, and $\lambda_\text{3d} = 1$. The piecewise-plane is estimated using the 10 nearest neighbors. We utilize the Adam optimizer to update both Gaussian parameters, with the learning rates for Gaussians identical to those used in the original Gaussian Splatting. All datasets are trained and evaluated for 30K iterations on a single NVIDIA 4090 GPU.

\subsection{Experimental Results}
\begin{figure*}[t!]
    \centering
    \includegraphics[width=1\linewidth]{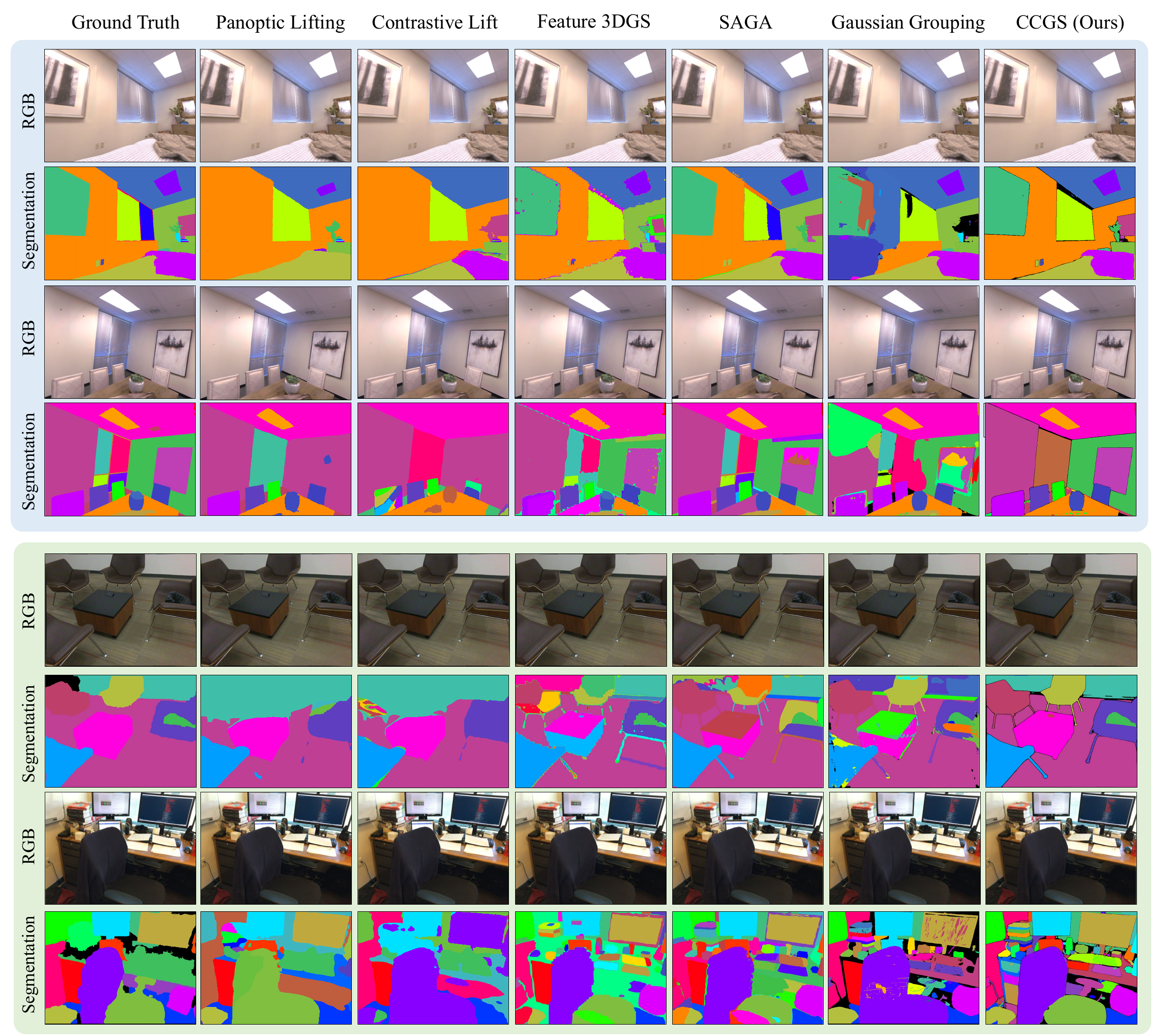}
    \caption{2D segmentation results on Replica and ScanNet datasets. Each column from left to right in the figure represents Ground truth segmentation, Panoptic Lifting, Contrastive Lift, Feature 3DGS, SAGA, Gaussian Grouping and Ours (CCGS). The top four lines represent different scenes in Replica. The following four lines are from different scenes in ScanNet.}
    \label{2D_vis}
\end{figure*}

\paragraph{2D segmentation} To validate CCGS on 2D panoptic segmentation task, we compare the results of Panoptic Lifting~\cite{siddiqui2023panoptic}, Contrastive Lift~\cite{bhalgat2023contrastive}, SAGA~\cite{hu2024semantic}, Feature 3DGS~\cite{zhou2024feature} and  Gaussian Grouping~\cite{ye2023gaussian} with CCGS on the Replica and ScanNet dataset. Since SAGA and Feature 3DGS are designed only for single-view segmentation tasks and cannot provide consistent segmentation IDs across multiple views, we match the segmentation IDs from their single-view results to the IDs generated by video segmentation methods~\cite{cheng2023tracking}. This serves as an extension of the video mask association approach.

As shown in Table~\ref{2D_data}, for single view $mIoU_{s}$, our CCGS method achieves better performance compared to Gaussian Grouping  with improvements of 2.38\% on the ScanNet dataset and 1.29\% on the Replica dataset. By comparing the single-view $mIoU_{s}$ and multi-view $mIoU_{m}$, our method experiences a 3\% drop in IoU, which is attributed to mismatches caused by differences in segmentation scales across views. Panoptic Lifting and Contrastive Lift also show a relatively small IoU drop of around 5-8\%, indicating that their methods maintain multi-view consistency. However, video mask association-based methods, such as Gaussian Grouping, Feature 3DGS, and SAGA, suffer a much larger IoU drop of about 20\%, which strongly suggests that video segmentation methods lack multi-view consistency in datasets with significant view differences. When compared to Gaussian Grouping on the Replica dataset, CCGS shows enhancements of 1.15 in PSNR, and 0.031 in SSIM. On the ScanNet dataset, CCGS similarly surpasses Gaussian Grouping by 1.05 in PSNR, and 0.052 in SSIM. The increase in PSNR and SSIM indicates that while achieving excellent 2D segmentation, the introduction of piecewise-plane  constraint also helps to better reconstruct the entire scene.

\begin{figure*}[h!]
    \centering
    \includegraphics[width=1\linewidth]{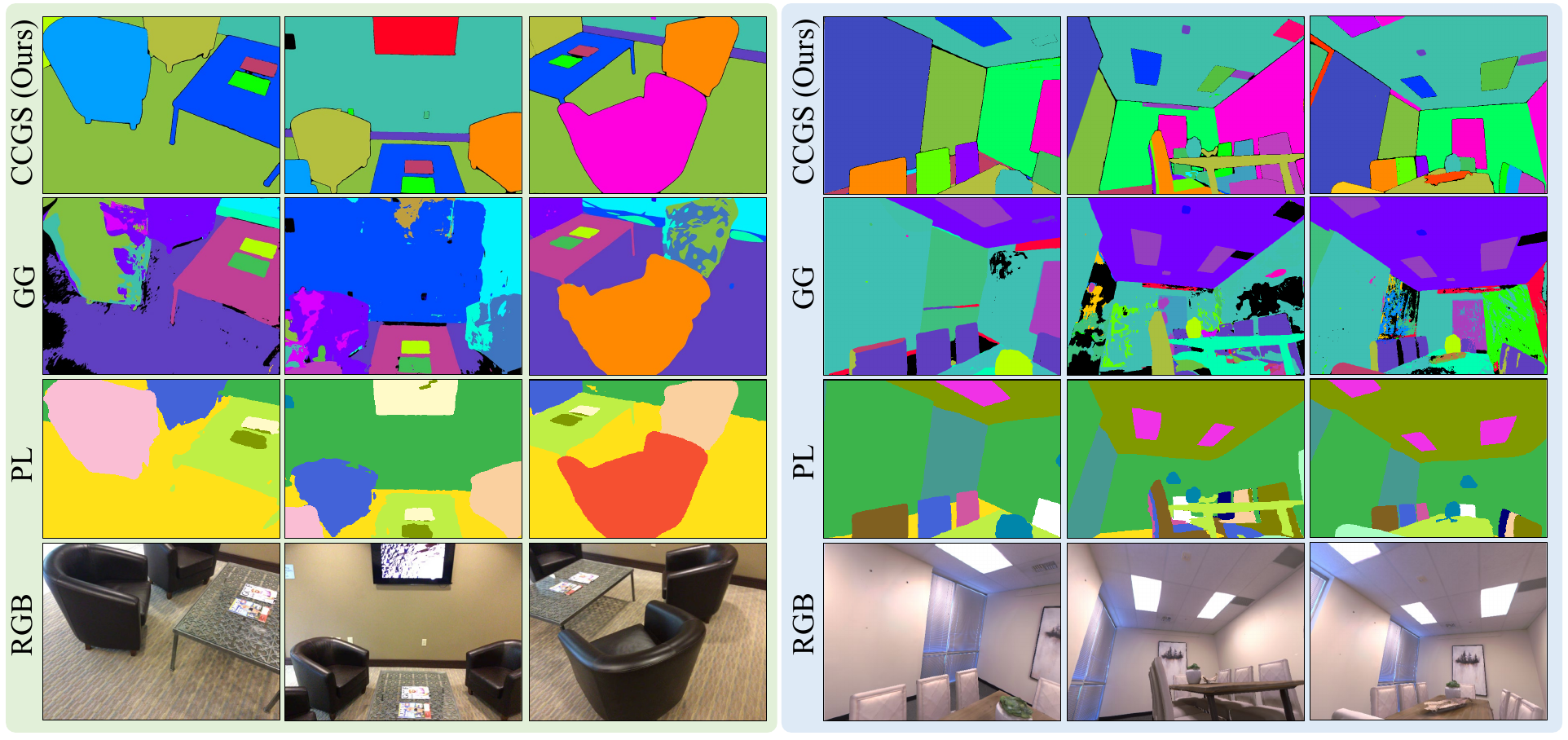}
    \caption{The comparison on multi-view consistency between CCGS and GG (Gaussian Grouping) and PL (Panoptic Lifting). From top to bottom, the images display CCGS, GG, PL and RGB inputs, respectively.}
    \label{views}
\end{figure*}

The qualitative results on the Replica and ScanNet datasets are presented in Figure~\ref{2D_vis}. Panoptic Lifting exhibits issues with incomplete segmentation and blurred boundaries. Contrastive Lift struggles to maintain training stability in complex scenarios, often leading to over-segmentation in the later stages of training. FeatureGS, on the other hand, suffers from feature confusion and uneven edges at object boundaries. This issue arises because FeatureGS only aligns SAM features with Gaussian point features at the 2D level without incorporating actual masks during training. SAGA demonstrates strong performance in 2D segmentation. However, it is inherently unsuitable for obtaining multi-view consistent masks. Due to the lack of spatial information in video segmentation method, Gaussian Grouping exhibits inconsistencies across multiple perspectives, leading to over-segmentation and numerous artifacts in the segmentation results during rendering. In contrast, CCGS, with pointmap fusion ensuring consistent segmentation, outperforms other methods in both segmentation completeness and accuracy.
\begin{table*}[ht!]
  \caption{The \textbf{2D panoptic segmentation} results on the Replica and ScanNet datasets demonstrate that CCGS outperforms other methods on both datasets.}
  \label{2D_data}
  \centering
  \begin{tabular}{l|cccc|cccc}
    \toprule
    &\multicolumn{4}{c|}{\textbf{Replica}} & \multicolumn{4}{c}{\textbf{ScanNet}} \\
    \textbf{Model} &$mIoU_{s}$   &$mIoU_{m}$  &PSNR  &SSIM   &$mIoU_{s}$  &$mIoU_{m}$   &PSNR &SSIM   \\
    \midrule
    Panoptic Lifting~\cite{siddiqui2023panoptic}&62.56&55.75
&30.23&0.892&57.24&52.33&26.23&0.812\\
    Contrastive Lift~\cite{bhalgat2023contrastive}&60.88&52.93
&30.26&0.901&55.30&49.61&26.18&0.813\\
    Feature 3DGS~\cite{zhou2024feature} & 63.84&43.51&32.31&0.926&58.08&40.37&27.25&0.882\\
    SAGA~\cite{hu2024semantic}&64.86&45.32&32.50&0.939&62.62&42.54&27.44
&0.890
\\
    Gaussian Grouping~\cite{ye2023gaussian} & 64.25&47.16
&32.41&0.935
&61.54&44.37&27.27&0.889\\
    CCGS (ours)& \textbf{65.54}&\textbf{62.31}& \textbf{33.56}&\textbf{0.966}&  \textbf{63.92}&\textbf{60.27}&\textbf{28.32}& \textbf{0.941}\\
    \bottomrule
  \end{tabular}
\end{table*}

Figure~\ref{views} presents the visualization results of multi-view consistency. While Panoptic Lifting achieves multi-view consistent segmentation, the quality of the segmentation boundaries remains relatively low. Gaussian Grouping tends to produce more artifacts when applied to datasets where objects are not centrally positioned. For instance, the same chair in Gaussian Grouping appears in different colors across various views, and the wall is over-segmented due to inconsistent IDs. In contrast, objects segmented by CCGS not only maintain consistent segmentation across different views but also preserve sharper boundaries and more coherent object structures.
\begin{figure*}[t]
    \centering
    \includegraphics[width=1\linewidth]{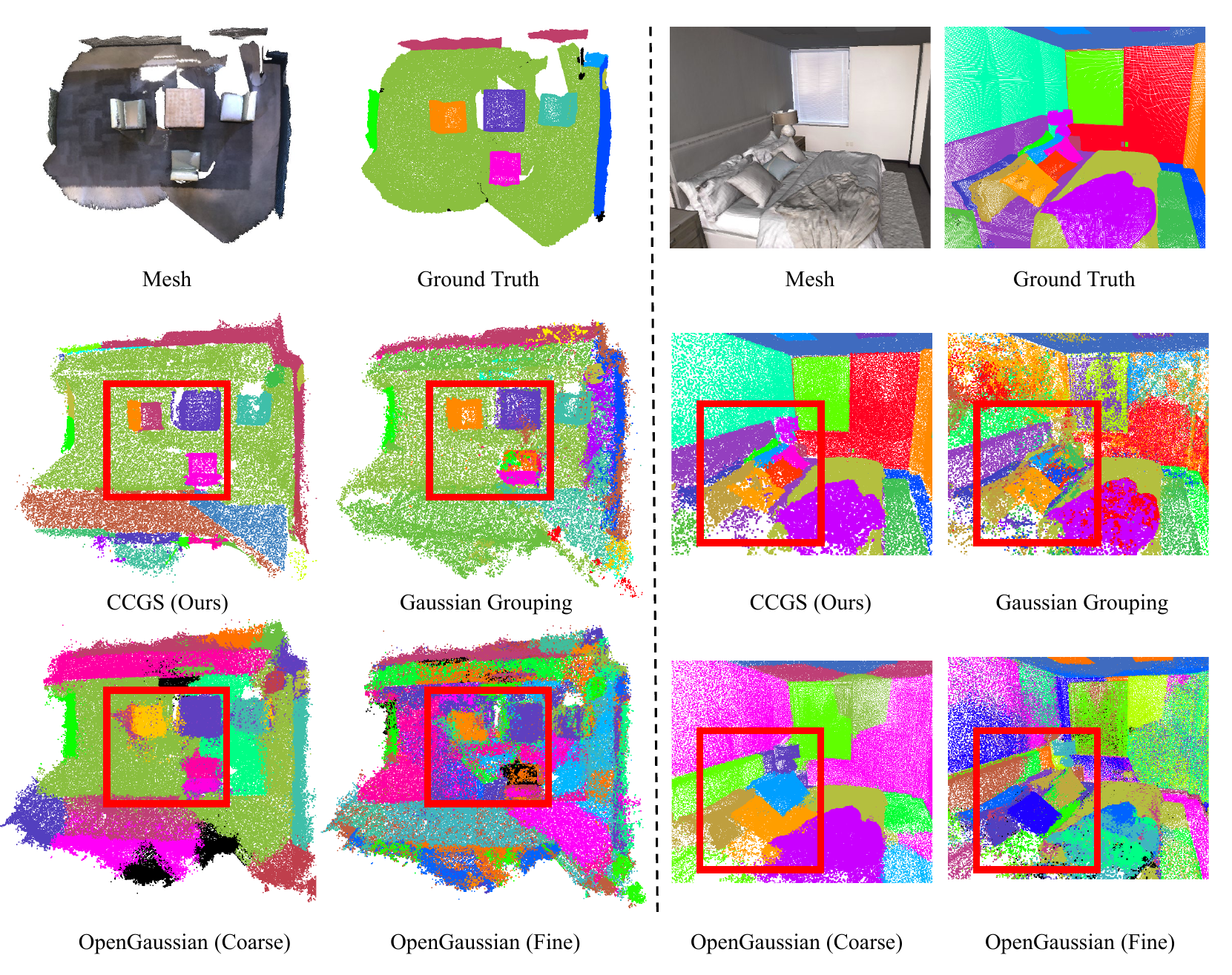}
    \caption{3D Gaussian segmentation results on ScanNet and Replica datasets. Each scene consists of a ground truth mesh, ground truth point cloud segmentation, our method (CCGS), Gaussian Grouping, as well as coarse-level and fine-level OpenGaussian results.}
    \label{3D_vis}
\end{figure*}
\paragraph{3D Gaussian segmentation} 

\begin{table*}[t!]
  \caption{\textbf{3D Gaussian segmentation} results on Replica and ScanNet datasets.}
  \label{3D_data}
  \centering
  \begin{tabular}{l|cc|cc}
    \toprule
    &\multicolumn{2}{c|}{Replica} & \multicolumn{2}{c}{ScanNet} \\
    {Model} &$mIoU_{3D}$ $\uparrow$ & Chamfer Distance $\downarrow$   &$mIoU_{3D}$ $\uparrow$  &Chamfer Distance $\downarrow$\\
    \midrule
    OpenGaussian(Fine)~\cite{wu2024opengaussian}&20.56&0.237&23.69&0.496\\
 OpenGaussian(Coarse)~\cite{wu2024opengaussian}&53.34&0.237&50.56&0.496\\
    Gaussian Grouping~\cite{ye2023gaussian} &54.12&0.230&52.04&0.482\\
    CCGS (ours)& \textbf{65.46}& \textbf{0.192}& \textbf{63.21}& \textbf{0.451}\\
    \bottomrule
  \end{tabular}
\end{table*}
To assess CCGS for the 3D Gaussian segmentation task, we compare its performance with that of Gaussian Grouping~\cite{ye2023gaussian} and OpenGaussian~\cite{wu2024opengaussian} on the Replica and ScanNet datasets. OpenGaussian focuses on point-level 3D understanding and proposes a two-stage codebook to discretize features from a coarse to a fine level. As detailed in Table~\ref{3D_data}, our CCGS method surpasses Gaussian Grouping on the Replica dataset with improvements of 11.34\% in $mIoU_{3D}$, and 0.038 in Chamfer Distance. Similarly, on the ScanNet dataset, CCGS exceeds Gaussian Grouping by 11.17\% in $mIoU_{3D}$, and  0.031 in Chamfer Distance. The difference between CCGS and GG in $mIoU_{3D}$ is significantly greater than in $mIoU_{s}$ in Table~\ref{2D_data}. This discrepancy arises because $mIoU_{s}$ does not account for multi-view consistency and only evaluates IoU for individual images and ground truth. At the coarse level, OpenGaussian and Gaussian Grouping achieve comparable $mIoU_{3D}$. However, at the fine level, $mIoU_{3D}$ of OpenGaussian is significantly lower. Meanwhile, the reduction in Chamfer Distance indicates that plane-constrained Gaussian splatting optimizes Gaussians with enhanced structural properties. 

\begin{figure*}[t!]
    \centering
    \includegraphics[width=1\linewidth]{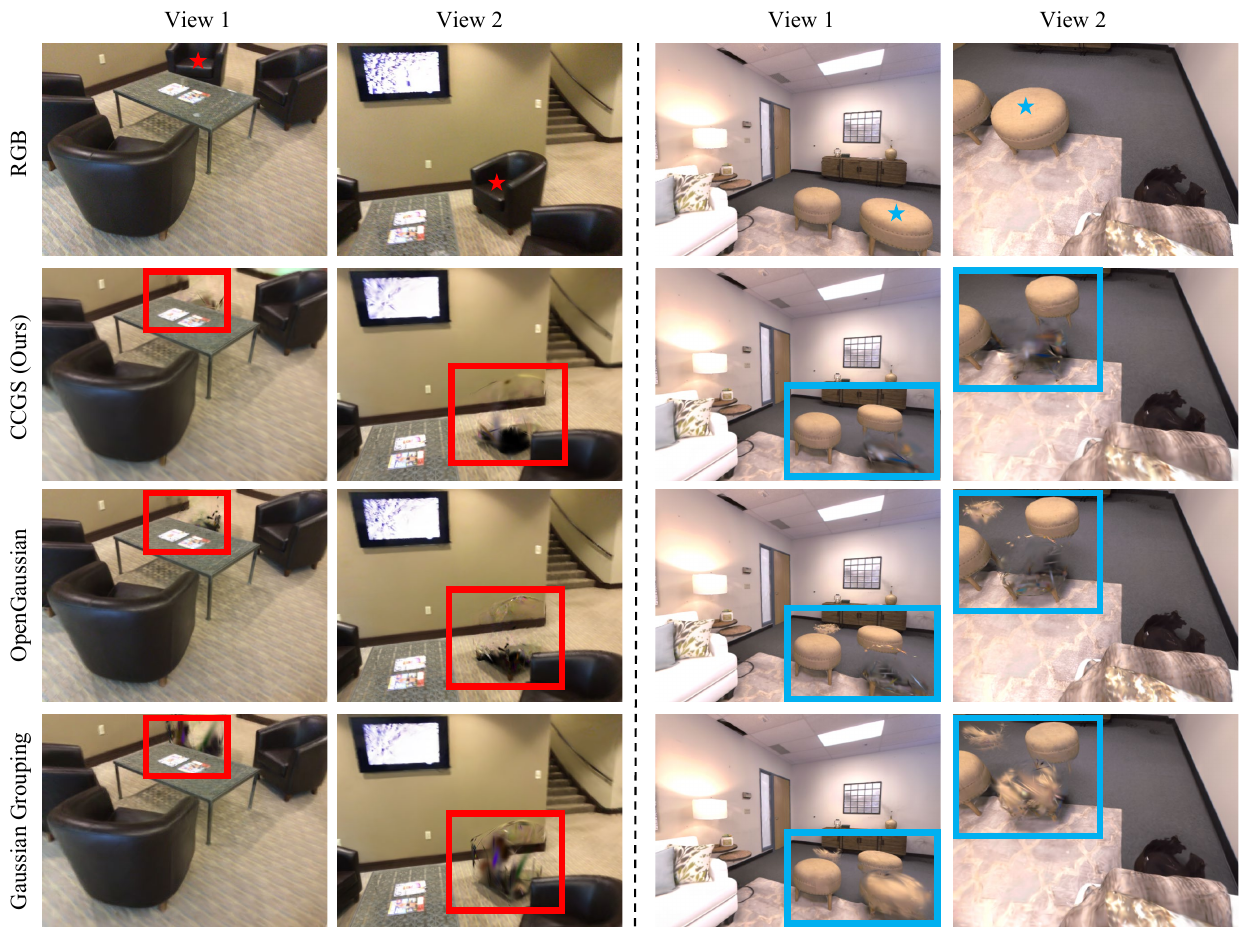}
    \caption{Results of the downstream deletion and movement tasks. The first row represents the RGB image, where $\textcolor{red}{\bigstar}$ marks the same objects to be deleted from different viewpoints, and $\textcolor[rgb]{0,0.690,0.941}{\bigstar}$ marks the same objects to be moved from different viewpoints.}
    \label{del}
\end{figure*}


As shown in the red box of the left scene in Figure~\ref{3D_vis}, both Gaussian Grouping and coarse-level OpenGaussian exhibit floaters. This issue is particularly evident in OpenGaussian, where yellow points originally belonging to the chair are propagated onto the purple coffee table, leading to classification boundary confusion and disrupting object integrity. In contrast, our CCGS method, leveraging plane regularization and split projection, effectively constrains the optimization of same category points within piecewise-planes. As a result, our approach achieves clear boundaries and well-structured reconstruction, significantly reducing floaters and improving spatial coherence. In the right-side scene, Gaussian Grouping exhibits noticeable color bleeding, where accumulated 2D inconsistencies lead to significant 3D inconsistencies. Coarse-level OpenGaussian fails to separate similar pillows, resulting in under-segmentation that merges distinct objects into a single cluster, while fine-level OpenGaussian suffers from over-segmentation. In contrast, our CCGS method achieves consistent segmentation with clear object boundaries.

\paragraph{Downstream tasks}

As shown in Figure~\ref{del}, in the deletion task, Gaussian grouping and OpenGaussian leaves many undeleted points due to inconsistent IDs. In contrast, our method effectively deletes Gaussian points, leaving minimal residue. In the movement task, both Gaussian Grouping and OpenGaussian cause parts of adjacent chairs to be moved as well. This occurs because the chairs are positioned closely together, and during optimization and cloning, category points from one chair are propagated to nearby similar chairs, leading to confusion between categories. In contrast, our method ensures that the movement of one chair does not affect other objects in the scene. Meanwhile, the ground after the movement is free from artifacts caused by residual points, unlike in Gaussian Grouping and OpenGaussian.

\begin{table*}[t!]
\caption{The comparison of different components PF (Pointmap Fusion), PR (Plane Regulaztion), SP (Split Projection) of CCGS on the ScanNet dataset.}
\centering
\renewcommand{\arraystretch}{0.8}
\begin{tabular}{ccc|ccccc}
\toprule
PF &  PR  & SP & $mIoU_{s}$~$\uparrow$& $mIoU_{m}$~$\uparrow$ & $mIoU_{3D}$~$\uparrow$ & Chamfer Distance~$\downarrow$\\
\midrule
-&-&-&61.54&44.37&52.04&0.482\\
\CheckmarkBold&-&-&63.38&60.18&57.12&0.480\\
\CheckmarkBold&\CheckmarkBold&-&63.83&60.21&59.56&0.469\\
\CheckmarkBold&\CheckmarkBold&\CheckmarkBold&\textbf{63.92}&\textbf{60.27}& \textbf{63.21}&\textbf{0.451}\\ 
\bottomrule
\end{tabular}
\label{tab:ablation}
\end{table*}
\subsection{Ablation Study}

\paragraph{Pointmap fusion} As shown in Table~\ref{tab:ablation}, incorporating pointmap fusion improves single-view $mIoU_{s}$ by 1.84\%, multi-view $mIoU_{m}$ by 15.81\%, $mIoU_{3D}$ by 5.08\%, and Chamfer Distance by 0.002. This demonstrates that pointmap fusion significantly enhances multi-view consistency, as reflected in the substantial improvements in $mIoU_{m}$ and $mIoU_{3D}$. These results highlight the superiority of pointmap fusion over traditional video segmentation methods in maintaining multi-view consistency. Meanwhile, the Chamfer Distance remains relatively unchanged, as pointmap fusion primarily aligns 2D segmentation pseudo-labels without modifying the underlying 3D Gaussian structure during training.

\paragraph{Plane regularization}
Plane regularization improves $mIoU_{s}$, $mIoU_{m}$, $mIoU_{3D}$, and Chamfer Distance by 0.45\%, 0.03\%, 2.44\%, and 0.011, respectively. Since we treat points that are far from the ground truth as unlabeled when calculating $mIoU_{3D}$, the compactness of the reconstruction directly contributes to the improvement in $mIoU_{3D}$. This demonstrates that plane regularization helps better preserve the movement of Gaussians within the plane of similar points, thereby enhancing the robustness of the reconstruction.

\paragraph{Split projection}
Split projection further enhances the performance of CCGS. As shown in Table~\ref{tab:ablation}, incorporating split projection improves $mIoU_{s}$, $mIoU_{m}$, $mIoU_{3D}$, and Chamfer Distance by 0.09\%, 0.06\%, 3.65\%, and 0.018, respectively. Compared to plane regularization, split projection is more effective in improving $mIoU_{3D}$. This suggests that the initialization of Gaussian positions during the split process introduces greater uncertainty in the segmentation of the 3D field. By projecting the initialized points onto the piecewise-plane, split projection enhances object compactness and reduces confusion at the boundaries.
\begin{figure}[t!]
    \centering
    \includegraphics[width=1\linewidth]{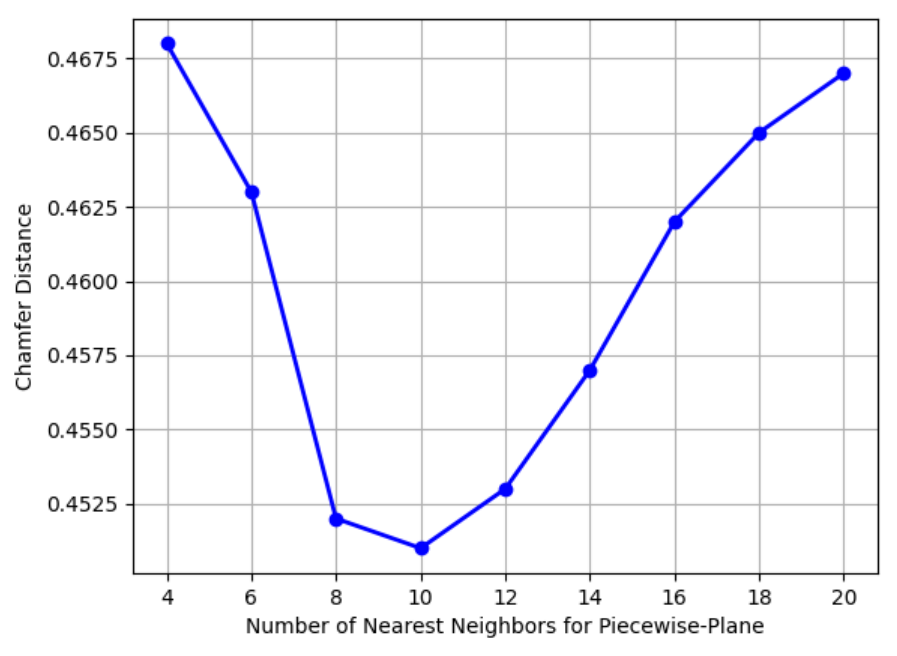}
    \caption{Chamfer Distance vs. Number of Nearest Neighbors for Piecewise-Plane}
    \label{neighbor}
\end{figure}
\paragraph{Nearest neighbors for piecewise-plane}
We conducted experiments on ScanNet, using 4 to 20 nearest neighbors to construct the piecewise-plane, selecting the optimal number based on Chamfer Distance. As shown in Figure~\ref{neighbor} , using only 4 points results in a higher Chamfer Distance. This is because a plane constructed with just 4 points cannot accurately represent the sub-plane in areas with dense Gaussian points. As the number of neighbors increases, Chamfer Distance decreases, reaching its minimum at 10 points. However, beyond this point, the distance starts to rise again. This happens because using too many neighbors causes regions that should be curved to become flattened, leading to deformation in some points.These findings highlight the importance of balancing local geometric fidelity and global structural integrity when constructing piecewise-planes for Gaussian optimization, ensuring accurate surface representation while preserving the overall scene structure.

\section{Limitations}  
Despite the promising results of CCGS, there are some limitations to consider. The quality of 2D segmentation may degrade in highly occluded or fast-moving viewpoints. While CCGS helps alleviate some of these issues, they can still negatively impact the overall segmentation performance. Future work could focus on optimizing 2D pseudo-labels during training, perhaps by incorporating more sophisticated temporal consistency constraints. Additionally, the current CCGS segmentation method is limited to static 3D scenes. Expanding the approach to dynamic 3D Gaussian segmentation, where objects can move and interact over time, will be an important direction for further research.

\section{Conclusion}  
We introduce CCGS, a consistent and compact 3D Gaussian segmentation field that significantly improves segmentation quality. By constructing a unified 3D field through pointmap fusion, CCGS effectively addresses inconsistencies caused by occlusions and viewpoint changes, ensuring reliable segmentation even in challenging scenarios. Additionally, we employ plane-constrained Gaussian splatting to ensure that points remain within their respective piecewise-planes, preventing the creation of ambiguous category points and improving segmentation clarity. Overall, our method substantially enhances the quality and consistency of segmentation results in both 2D and 3D domains, offering promising applications in a wide range of scene understanding, manipulation, and editing tasks.

\backmatter




\bibliography{ref}

\begin{thebibliography}{10}

\bibitem{bhalgat2023contrastive}
Y.~Bhalgat, I.~Laina, J.~F. Henriques, A.~Zisserman, and A.~Vedaldi.
\newblock Contrastive lift: 3d object instance segmentation by slow-fast contrastive fusion.
\newblock {\em arXiv preprint arXiv:2306.04633}, 2023.

\bibitem{bian2024dynamiccity}
H.~Bian, L.~Kong, H.~Xie, L.~Pan, Y.~Qiao, and Z.~Liu.
\newblock Dynamiccity: Large-scale lidar generation from dynamic scenes.
\newblock {\em arXiv preprint arXiv:2410.18084}, 2024.

\bibitem{castillo2024contrastive}
M.~Castillo, M.~Dahaghin, M.~Toso, and A.~Del~Bue.
\newblock Contrastive gaussian clustering for weakly supervised 3d scene segmentation.
\newblock In {\em International Conference on Pattern Recognition}, pages 114--130. Springer, 2024.

\bibitem{cen2023segment}
J.~Cen, J.~Fang, C.~Yang, L.~Xie, X.~Zhang, W.~Shen, and Q.~Tian.
\newblock Segment any 3d gaussians.
\newblock {\em arXiv preprint arXiv:2312.00860}, 2023.

\bibitem{chen2024ovgaussian}
R.~Chen, X.~Sun, Z.~Wang, Y.~Liu, J.~Wang, L.~Kong, J.~Deng, M.~Gong, L.~Pan, W.~Wang, et~al.
\newblock Ovgaussian: Generalizable 3d gaussian segmentation with open vocabularies.
\newblock {\em arXiv preprint arXiv:2501.00326}, 2024.

\bibitem{cheng2023tracking}
H.~K. Cheng, S.~W. Oh, B.~Price, A.~Schwing, and J.-Y. Lee.
\newblock Tracking anything with decoupled video segmentation.
\newblock In {\em Proceedings of the IEEE/CVF International Conference on Computer Vision}, pages 1316--1326, 2023.

\bibitem{cheng2024occam}
J.~Cheng, J.-N. Zaech, L.~Van~Gool, and D.~P. Paudel.
\newblock Occam's lgs: A simple approach for language gaussian splatting.
\newblock {\em arXiv preprint arXiv:2412.01807}, 2024.

\bibitem{dai2017scannet}
A.~Dai, A.~X. Chang, M.~Savva, M.~Halber, T.~Funkhouser, and M.~Nie{\ss}ner.
\newblock Scannet: Richly-annotated 3d reconstructions of indoor scenes.
\newblock In {\em Proceedings of the IEEE conference on computer vision and pattern recognition}, pages 5828--5839, 2017.

\bibitem{dou2024cosseggaussians}
B.~Dou, T.~Zhang, Y.~Ma, Z.~Wang, and Z.~Yuan.
\newblock Cosseggaussians: Compact and swift scene segmenting 3d gaussians.
\newblock {\em arXiv preprint arXiv:2401.05925}, 2024.

\bibitem{fu2022panoptic}
X.~Fu, S.~Zhang, T.~Chen, Y.~Lu, L.~Zhu, X.~Zhou, A.~Geiger, and Y.~Liao.
\newblock Panoptic nerf: 3d-to-2d label transfer for panoptic urban scene segmentation.
\newblock In {\em 2022 International Conference on 3D Vision (3DV)}, pages 1--11. IEEE, 2022.

\bibitem{goel2023interactive}
R.~Goel, D.~Sirikonda, S.~Saini, and P.~Narayanan.
\newblock Interactive segmentation of radiance fields.
\newblock In {\em Proceedings of the IEEE/CVF Conference on Computer Vision and Pattern Recognition}, pages 4201--4211, 2023.

\bibitem{hu2024sparselgs}
J.~Hu, Z.~Chen, Z.~Li, Y.~Xu, and J.~Zhang.
\newblock Sparselgs: Sparse view language embedded gaussian splatting.
\newblock {\em arXiv preprint arXiv:2412.02245}, 2024.

\bibitem{hu2024semantic}
X.~Hu, Y.~Wang, L.~Fan, J.~Fan, J.~Peng, Z.~Lei, Q.~Li, and Z.~Zhang.
\newblock Semantic anything in 3d gaussians.
\newblock {\em arXiv preprint arXiv:2401.17857}, 2024.

\bibitem{huang2024segment3d}
R.~Huang, S.~Peng, A.~Takmaz, F.~Tombari, M.~Pollefeys, S.~Song, G.~Huang, and F.~Engelmann.
\newblock Segment3d: Learning fine-grained class-agnostic 3d segmentation without manual labels.
\newblock In {\em European Conference on Computer Vision}, pages 278--295. Springer, 2024.

\bibitem{jain2024gaussiancut}
U.~Jain, A.~Mirzaei, and I.~Gilitschenski.
\newblock Gaussiancut: Interactive segmentation via graph cut for 3d gaussian splatting.
\newblock In {\em The Thirty-eighth Annual Conference on Neural Information Processing Systems}, 2024.

\bibitem{jaritz2019multi}
M.~Jaritz, J.~Gu, and H.~Su.
\newblock Multi-view pointnet for 3d scene understanding.
\newblock In {\em Proceedings of the IEEE/CVF international conference on computer vision workshops}, pages 0--0, 2019.

\bibitem{kamran2024applications}
A.~Kamran-Pishhesari, A.~Moniri-Morad, and J.~Sattarvand.
\newblock Applications of 3d reconstruction in virtual reality-based teleoperation: A review in the mining industry.
\newblock {\em Technologies}, 12(3):40, 2024.

\bibitem{kerbl20233d}
B.~Kerbl, G.~Kopanas, T.~Leimk{\"u}hler, and G.~Drettakis.
\newblock 3d gaussian splatting for real-time radiance field rendering.
\newblock {\em ACM Transactions on Graphics}, 42(4):1--14, 2023.

\bibitem{kerr2023lerf}
J.~Kerr, C.~M. Kim, K.~Goldberg, A.~Kanazawa, and M.~Tancik.
\newblock Lerf: Language embedded radiance fields.
\newblock In {\em Proceedings of the IEEE/CVF International Conference on Computer Vision}, pages 19729--19739, 2023.

\bibitem{kim2024garfield}
C.~M. Kim, M.~Wu, J.~Kerr, K.~Goldberg, M.~Tancik, and A.~Kanazawa.
\newblock Garfield: Group anything with radiance fields.
\newblock In {\em Proceedings of the IEEE/CVF Conference on Computer Vision and Pattern Recognition}, pages 21530--21539, 2024.

\bibitem{kirillov2023segment}
A.~Kirillov, E.~Mintun, N.~Ravi, H.~Mao, C.~Rolland, L.~Gustafson, T.~Xiao, S.~Whitehead, A.~C. Berg, W.-Y. Lo, et~al.
\newblock Segment anything.
\newblock In {\em Proceedings of the IEEE/CVF International Conference on Computer Vision}, pages 4015--4026, 2023.

\bibitem{kong2023robo3d}
L.~Kong, Y.~Liu, X.~Li, R.~Chen, W.~Zhang, J.~Ren, L.~Pan, K.~Chen, and Z.~Liu.
\newblock Robo3d: Towards robust and reliable 3d perception against corruptions.
\newblock In {\em Proceedings of the IEEE/CVF International Conference on Computer Vision}, pages 19994--20006, 2023.

\bibitem{kong2024multi}
L.~Kong, X.~Xu, J.~Ren, W.~Zhang, L.~Pan, K.~Chen, W.~T. Ooi, and Z.~Liu.
\newblock Multi-modal data-efficient 3d scene understanding for autonomous driving.
\newblock {\em arXiv preprint arXiv:2405.05258}, 2024.

\bibitem{kundu2022panoptic}
A.~Kundu, K.~Genova, X.~Yin, A.~Fathi, C.~Pantofaru, L.~J. Guibas, A.~Tagliasacchi, F.~Dellaert, and T.~Funkhouser.
\newblock Panoptic neural fields: A semantic object-aware neural scene representation.
\newblock In {\em Proceedings of the IEEE/CVF Conference on Computer Vision and Pattern Recognition}, pages 12871--12881, 2022.

\bibitem{kundu2020virtual}
A.~Kundu, X.~Yin, A.~Fathi, D.~Ross, B.~Brewington, T.~Funkhouser, and C.~Pantofaru.
\newblock Virtual multi-view fusion for 3d semantic segmentation.
\newblock In {\em Computer Vision--ECCV 2020: 16th European Conference, Glasgow, UK, August 23--28, 2020, Proceedings, Part XXIV 16}, pages 518--535. Springer, 2020.

\bibitem{li2024instancegaussian}
H.~Li, Y.~Wu, J.~Meng, Q.~Gao, Z.~Zhang, R.~Wang, and J.~Zhang.
\newblock Instancegaussian: Appearance-semantic joint gaussian representation for 3d instance-level perception.
\newblock {\em arXiv preprint arXiv:2411.19235}, 2024.

\bibitem{li2024sadg}
Y.-J. Li, M.~Gladkova, Y.~Xia, and D.~Cremers.
\newblock Sadg: Segment any dynamic gaussian without object trackers.
\newblock {\em arXiv preprint arXiv:2411.19290}, 2024.

\bibitem{li2024gradiseg}
Z.~Li, W.~Han, Y.~Cai, H.~Jiang, B.~Bi, S.~Gao, H.~Zhao, and Z.~Wang.
\newblock Gradiseg: Gradient-guided gaussian segmentation with enhanced 3d boundary precision.
\newblock {\em arXiv preprint arXiv:2412.00392}, 2024.

\bibitem{liang2024supergseg}
S.~Liang, S.~Wang, K.~Li, M.~Niemeyer, S.~Gasperini, N.~Navab, and F.~Tombari.
\newblock Supergseg: Open-vocabulary 3d segmentation with structured super-gaussians.
\newblock {\em arXiv preprint arXiv:2412.10231}, 2024.

\bibitem{liu2024active}
G.~Liu, O.~van Kaick, H.~Huang, and R.~Hu.
\newblock Active self-training for weakly supervised 3d scene semantic segmentation.
\newblock {\em Computational Visual Media}, 10(3):425--438, 2024.

\bibitem{liu2023weakly}
K.~Liu, F.~Zhan, J.~Zhang, M.~Xu, Y.~Yu, A.~El~Saddik, C.~Theobalt, E.~Xing, and S.~Lu.
\newblock Weakly supervised 3d open-vocabulary segmentation.
\newblock {\em Advances in Neural Information Processing Systems}, 36:53433--53456, 2023.

\bibitem{lyu2024gaga}
W.~Lyu, X.~Li, A.~Kundu, Y.-H. Tsai, and M.-H. Yang.
\newblock Gaga: Group any gaussians via 3d-aware memory bank.
\newblock {\em arXiv preprint arXiv:2404.07977}, 2024.

\bibitem{mildenhall2021nerf}
B.~Mildenhall, P.~P. Srinivasan, M.~Tancik, J.~T. Barron, R.~Ramamoorthi, and R.~Ng.
\newblock Nerf: Representing scenes as neural radiance fields for view synthesis.
\newblock {\em Communications of the ACM}, 65(1):99--106, 2021.

\bibitem{park2022fast}
C.~Park, Y.~Jeong, M.~Cho, and J.~Park.
\newblock Fast point transformer.
\newblock In {\em Proceedings of the IEEE/CVF Conference on Computer Vision and Pattern Recognition}, pages 16949--16958, 2022.

\bibitem{peng20243d}
Q.~Peng, B.~Planche, Z.~Gao, M.~Zheng, A.~Choudhuri, T.~Chen, C.~Chen, and Z.~Wu.
\newblock 3d vision-language gaussian splatting.
\newblock {\em arXiv preprint arXiv:2410.07577}, 2024.

\bibitem{peng2023openscene}
S.~Peng, K.~Genova, C.~Jiang, A.~Tagliasacchi, M.~Pollefeys, T.~Funkhouser, et~al.
\newblock Openscene: 3d scene understanding with open vocabularies.
\newblock In {\em Proceedings of the IEEE/CVF Conference on Computer Vision and Pattern Recognition}, pages 815--824, 2023.

\bibitem{peng2024gags}
Y.~Peng, H.~Wang, Y.~Liu, C.~Wen, Z.~Dong, and B.~Yang.
\newblock Gags: Granularity-aware feature distillation for language gaussian splatting.
\newblock {\em arXiv preprint arXiv:2412.13654}, 2024.

\bibitem{qi2017pointnet}
C.~R. Qi, H.~Su, K.~Mo, and L.~J. Guibas.
\newblock Pointnet: Deep learning on point sets for 3d classification and segmentation.
\newblock In {\em Proceedings of the IEEE conference on computer vision and pattern recognition}, pages 652--660, 2017.

\bibitem{qi2017pointnet++}
C.~R. Qi, L.~Yi, H.~Su, and L.~J. Guibas.
\newblock Pointnet++: Deep hierarchical feature learning on point sets in a metric space.
\newblock {\em Advances in neural information processing systems}, 30, 2017.

\bibitem{qin2024langsplat}
M.~Qin, W.~Li, J.~Zhou, H.~Wang, and H.~Pfister.
\newblock Langsplat: 3d language gaussian splatting.
\newblock In {\em Proceedings of the IEEE/CVF Conference on Computer Vision and Pattern Recognition}, pages 20051--20060, 2024.

\bibitem{qiu2024gls}
J.~Qiu, L.~Liu, Z.~Su, and T.~Lin.
\newblock Gls: Geometry-aware 3d language gaussian splatting.
\newblock {\em arXiv preprint arXiv:2411.18066}, 2024.

\bibitem{radford2021learning}
A.~Radford, J.~W. Kim, C.~Hallacy, A.~Ramesh, G.~Goh, S.~Agarwal, G.~Sastry, A.~Askell, P.~Mishkin, J.~Clark, et~al.
\newblock Learning transferable visual models from natural language supervision.
\newblock In {\em International conference on machine learning}, pages 8748--8763. PmLR, 2021.

\bibitem{rozenberszki2024unscene3d}
D.~Rozenberszki, O.~Litany, and A.~Dai.
\newblock Unscene3d: Unsupervised 3d instance segmentation for indoor scenes.
\newblock In {\em Proceedings of the IEEE/CVF Conference on Computer Vision and Pattern Recognition}, pages 19957--19967, 2024.

\bibitem{shen2023distilled}
W.~Shen, G.~Yang, A.~Yu, J.~Wong, L.~P. Kaelbling, and P.~Isola.
\newblock Distilled feature fields enable few-shot language-guided manipulation.
\newblock {\em arXiv preprint arXiv:2308.07931}, 2023.

\bibitem{shi2024language}
J.-C. Shi, M.~Wang, H.-B. Duan, and S.-H. Guan.
\newblock Language embedded 3d gaussians for open-vocabulary scene understanding.
\newblock In {\em Proceedings of the IEEE/CVF Conference on Computer Vision and Pattern Recognition}, pages 5333--5343, 2024.

\bibitem{siddiqui2023panoptic}
Y.~Siddiqui, L.~Porzi, S.~R. Bul{\'o}, N.~M{\"u}ller, M.~Nie{\ss}ner, A.~Dai, and P.~Kontschieder.
\newblock Panoptic lifting for 3d scene understanding with neural fields.
\newblock In {\em Proceedings of the IEEE/CVF Conference on Computer Vision and Pattern Recognition}, pages 9043--9052, 2023.

\bibitem{straub2019replica}
J.~Straub, T.~Whelan, L.~Ma, Y.~Chen, E.~Wijmans, S.~Green, J.~J. Engel, R.~Mur-Artal, C.~Ren, S.~Verma, et~al.
\newblock The replica dataset: A digital replica of indoor spaces.
\newblock {\em arXiv preprint arXiv:1906.05797}, 2019.

\bibitem{tang2022contrastive}
L.~Tang, Y.~Zhan, Z.~Chen, B.~Yu, and D.~Tao.
\newblock Contrastive boundary learning for point cloud segmentation.
\newblock In {\em Proceedings of the IEEE/CVF Conference on Computer Vision and Pattern Recognition}, pages 8489--8499, 2022.

\bibitem{wang2022dm}
B.~Wang, L.~Chen, and B.~Yang.
\newblock Dm-nerf: 3d scene geometry decomposition and manipulation from 2d images.
\newblock {\em arXiv preprint arXiv:2208.07227}, 2022.

\bibitem{wang2023dust3r}
S.~Wang, V.~Leroy, Y.~Cabon, B.~Chidlovskii, and J.~Revaud.
\newblock Dust3r: Geometric 3d vision made easy.
\newblock {\em arXiv preprint arXiv:2312.14132}, 2023.

\bibitem{wang2024plgs}
Y.~Wang, X.~Wei, M.~Lu, and G.~Kang.
\newblock Plgs: Robust panoptic lifting with 3d gaussian splatting.
\newblock {\em arXiv preprint arXiv:2410.17505}, 2024.

\bibitem{wei2024nto3d}
X.~Wei, R.~Zhang, J.~Wu, J.~Liu, M.~Lu, Y.~Guo, and S.~Zhang.
\newblock Nto3d: Neural target object 3d reconstruction with segment anything.
\newblock In {\em Proceedings of the IEEE/CVF Conference on Computer Vision and Pattern Recognition}, pages 20352--20362, 2024.

\bibitem{wu2024opengaussian}
Y.~Wu, J.~Meng, H.~Li, C.~Wu, Y.~Shi, X.~Cheng, C.~Zhao, H.~Feng, E.~Ding, J.~Wang, et~al.
\newblock Opengaussian: Towards point-level 3d gaussian-based open vocabulary understanding.
\newblock {\em arXiv preprint arXiv:2406.02058}, 2024.

\bibitem{xia2023scpnet}
Z.~Xia, Y.~Liu, X.~Li, X.~Zhu, Y.~Ma, Y.~Li, Y.~Hou, and Y.~Qiao.
\newblock Scpnet: Semantic scene completion on point cloud.
\newblock In {\em Proceedings of the IEEE/CVF conference on computer vision and pattern recognition}, pages 17642--17651, 2023.

\bibitem{xu2024embodiedsam}
X.~Xu, H.~Chen, L.~Zhao, Z.~Wang, J.~Zhou, and J.~Lu.
\newblock Embodiedsam: Online segment any 3d thing in real time.
\newblock {\em arXiv preprint arXiv:2408.11811}, 2024.

\bibitem{ye2023gaussian}
M.~Ye, M.~Danelljan, F.~Yu, and L.~Ke.
\newblock Gaussian grouping: Segment and edit anything in 3d scenes.
\newblock {\em arXiv preprint arXiv:2312.00732}, 2023.

\bibitem{zanjani2024planar}
F.~G. Zanjani, H.~Cai, H.~Ackermann, L.~Mirvakhabova, and F.~Porikli.
\newblock Planar gaussian splatting.
\newblock {\em arXiv preprint arXiv:2412.01931}, 2024.

\bibitem{zhang2024bootstraping}
W.~Zhang, L.~Zhang, P.~Hu, L.~Ma, Y.~Zhuge, and H.~Lu.
\newblock Bootstraping clustering of gaussians for view-consistent 3d scene understanding.
\newblock {\em arXiv preprint arXiv:2411.19551}, 2024.

\bibitem{zhi2021place}
S.~Zhi, T.~Laidlow, S.~Leutenegger, and A.~J. Davison.
\newblock In-place scene labelling and understanding with implicit scene representation.
\newblock In {\em Proceedings of the IEEE/CVF International Conference on Computer Vision}, pages 15838--15847, 2021.

\bibitem{zhou2024feature}
S.~Zhou, H.~Chang, S.~Jiang, Z.~Fan, Z.~Zhu, D.~Xu, P.~Chari, S.~You, Z.~Wang, and A.~Kadambi.
\newblock Feature 3dgs: Supercharging 3d gaussian splatting to enable distilled feature fields.
\newblock In {\em Proceedings of the IEEE/CVF Conference on Computer Vision and Pattern Recognition}, pages 21676--21685, 2024.

\end{thebibliography}
\bibliographystyle{abbrv}

\end{document}